\documentclass[10pt,twocolumn,letterpaper]{article}

\usepackage{cvpr}              
\definecolor{cvprblue}{rgb}{0.21,0.49,0.74}
\usepackage[pagebackref,breaklinks,colorlinks,allcolors=cvprblue]{hyperref}

\usepackage{makecell}
\usepackage{amssymb}
\usepackage{pifont}
\usepackage[table]{xcolor}
\usepackage{multirow}
\usepackage{amsmath}
\usepackage{amsfonts}
\usepackage{mathrsfs}
\usepackage{booktabs} 
\definecolor{popo_red}{rgb}{0.75,0,0}
\definecolor{popo_green}{rgb}{0.47,0.58,0.235}
\usepackage{tcolorbox}

\def\paperID{*****} 
\def\confName{CVPR}
\def\confYear{2026}

\title{Venus: Benchmarking and Empowering Multimodal Large Language Models for Aesthetic Guidance and Cropping}

\author{Tianxiang Du, Hulingxiao He, Yuxin Peng$^*$\\
Wangxuan Institute of Computer Technology, Peking University \\
{\tt\small txdu25@stu.pku.edu.cn, hehulingxiao@stu.pku.edu.cn, pengyuxin@pku.edu.cn}}

\let\oldtwocolumn\twocolumn
\renewcommand\twocolumn[1][]{%
    \oldtwocolumn[{#1}{
    \begin{center}
           \includegraphics[width=\textwidth]{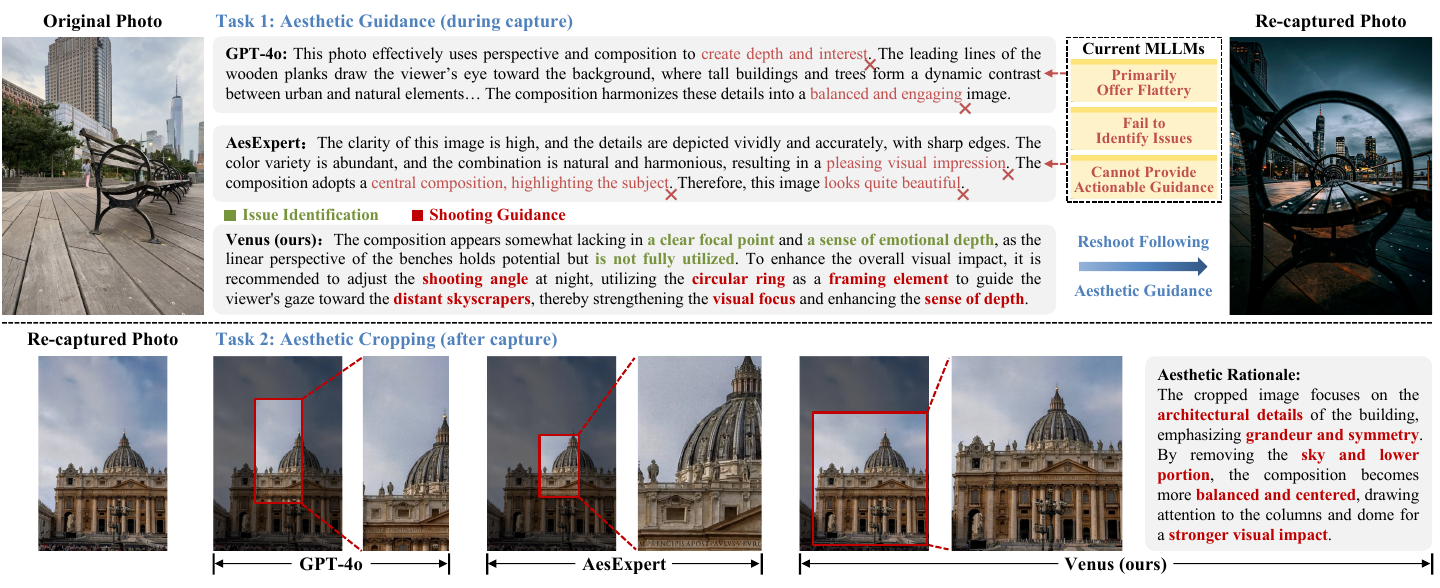}
           \captionof{figure}{Comparison of Venus (ours), GPT-4o (a general MLLM), and AesExpert (an aesthetic-oriented MLLM) across two tasks: \textbf{(1) Aesthetic Guidance (during capture)} and \textbf{(2) Aesthetic Cropping (after capture)}. In Task 1, existing MLLMs primarily offer flattery, fail to identify issues, and cannot provide actionable guidance. In Task 2, Venus produces balanced and visually appealing crops with clear aesthetic rationales, whereas existing MLLMs fail to reframe effectively or offer convincing explanations.} 
           \label{Fig.1}
        \end{center}
    }]
}

\begin{document}
\maketitle
\insert\footins{\footnotesize$^*$Corresponding author.}
\begin{abstract}
The widespread use of smartphones has made photography ubiquitous, yet a clear gap remains between ordinary users and professional photographers, who can identify aesthetic issues and provide actionable shooting guidance during capture. We define this capability as \textbf{aesthetic guidance (AG)} --- an essential but largely underexplored domain in computational aesthetics. Existing multimodal large language models (MLLMs) primarily offer overly positive feedback, failing to identify issues or provide actionable guidance. Without AG capability, they cannot effectively identify distracting regions or optimize compositional balance, thus also struggling in \textbf{aesthetic cropping}, which aims to refine photo composition through reframing after capture. To address this, we introduce \textbf{AesGuide}, the first large-scale AG dataset and benchmark with 10,748 photos annotated with aesthetic scores, analyses, and guidance. Building upon it, we propose \textbf{Venus}, a two-stage framework that first empowers MLLMs with AG capability through progressively complex aesthetic questions and then activates their aesthetic cropping power via CoT-based rationales. Extensive experiments show that Venus substantially improves AG capability and achieves state-of-the-art (SOTA) performance in aesthetic cropping, enabling interpretable and interactive aesthetic refinement across both stages of photo creation. Code is available at \href{https://github.com/PKU-ICST-MIPL/Venus_CVPR2026}{https://github.com/PKU-ICST-MIPL/Venus\_CVPR2026}.
\end{abstract}    
\section{Introduction}
\label{sec:introduction}

Smartphones have made photography an integral and common part of daily life~\cite{lou2021Jin, zhao2025PICD, liao2025thinking}. However, without professional photographic expertise, most users struggle to capture aesthetically pleasing photos~\cite{jiang2022photohelper, liu2025CALM, li2025SPAS}. In comparison, professional photographers can identify aesthetic issues (\eg, unbalanced composition or improper lighting) and provide actionable shooting guidance (\eg, adjusting viewpoint or modifying lighting) during capture to achieve superior visual results. We define this capability as \textbf{aesthetic guidance (AG)} --- an essential but largely underexplored domain in computational aesthetics, which bridges the gap between subjective aesthetic understanding and objective, actionable shooting adjustments.

Multimodal large language models (MLLMs)~\cite{liu2023llava, bai2023qwen-vl-chat, chen2024internVL2.5, achiam2023gpt-4, bai2025qwen25, peng2025CJE_survey} have achieved remarkable progress on general tasks~\cite{zhang2024CJE_fscil,jiang2024CJE_towards, yang2025CJE_gala}. However, as shown in \cref{Fig.1}, their performance in AG remains limited. Both general MLLMs (\eg, GPT-4o~\cite{hurst2024gpt-4o}) and aesthetic-oriented ones (\eg, AesExpert~\cite{huang2024Aesexpert}) tend to produce overly positive feedback~\cite{qi2025PhotoEye, cao2025artimuse}, failing to identify aesthetic flaws or provide actionable guidance. Lacking AG capability, they cannot effectively identify distracting regions or optimize compositional balance, thereby also struggling in \textbf{aesthetic cropping}, which aims to refine photo composition through reframing after capture. 

On the other hand, existing specialized aesthetic cropping models~\cite{hong2021CACNet, yang2023SAC-Net, hong2024Gencrop, su2024UGC, zhang2025procrop} are typically small-scale, architecturally diverse, and narrowly limited to the cropping task. More critically, they lack interpretability and interactivity, preventing them from explaining the underlying aesthetic rationales behind their crops or dynamically adapting to user preferences. Given the inherently subjective nature of aesthetics and users’ diverse compositional tastes, such models fail to meet varied aesthetic demands.

\begin{figure*}[htbp]
\centering
\includegraphics[width=\textwidth]{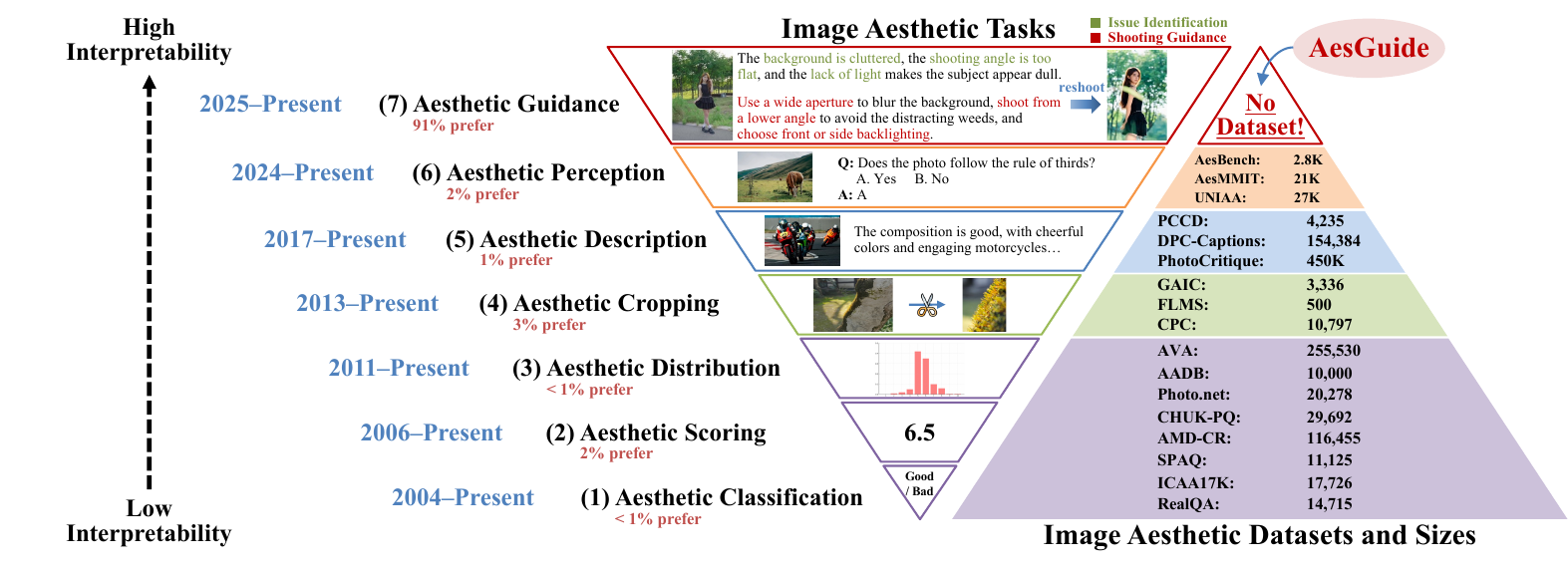}
\caption{Overview of image aesthetic tasks and datasets. We follow and refine the comprehensive aesthetic task taxonomy proposed by Jin \etal~\cite{jin2024apddv2}. A user survey of 1,069 participants shows that 91\% prefer AG, a largely underexplored task with no dedicated dataset.}
\label{Fig.2}
\end{figure*}

These challenges arise from both dataset and method gaps. As shown in \cref{Fig.2}, existing aesthetic datasets for MLLMs~\cite{chang2017PCCD, zhou2024uniaa, huang2024aesbench, huang2024Aesexpert, qi2025PhotoEye} mainly target aesthetic perception and description tasks, with evaluations that tend to be overly positive~\cite{zhou2024uniaa, huang2024aesbench}, thereby limiting their real-world applicability. Meanwhile, current MLLMs remain misaligned with human aesthetic reasoning, preventing them from generating reliable and actionable guidance. A user survey of 1,069 participants further reveals that 91\% prefer AG functionality, highlighting a strong demand that existing datasets and methods still fail to meet.

To address these challenges, we introduce \textbf{AesGuide}, the first AG dataset and benchmark annotated with aesthetic scores, analyses, and guidance, along with three evaluation metrics and expert assessments across 10,748 real-world photos.
Building on it, we propose \textbf{Venus}, a two-stage framework comprising:
\textbf{(1) Aesthetic guidance capability building}, which empowers MLLMs with AG capability through progressively complex aesthetic questions that evolve from overall impressions to detailed analyses and guidance, thereby aligning with human aesthetic reasoning.
\textbf{(2) Aesthetic cropping power activation}, which unlocks cropping ability via Chain-of-Thought (CoT) rationales that convey the aesthetic logic behind cropping decisions.
Extensive experiments demonstrate that Venus substantially improves AG performance across five open-source MLLMs and achieves state-of-the-art (SOTA) results in aesthetic cropping, enabling interpretable and interactive aesthetic refinement across both stages of photo creation.

The main contributions are summarized as follows:
\begin{itemize}
\item[$\bullet$] We formally define \textbf{aesthetic guidance (AG)}, an underexplored domain bridging subjective aesthetic understanding with objective, actionable shooting adjustments.
\item[$\bullet$] \textbf{AesGuide} is presented as the first AG benchmark, annotated with aesthetic scores, analyses, and guidance, along with three evaluation metrics and expert assessments.
\item[$\bullet$] We propose \textbf{Venus}, a two-stage framework that first empowers MLLMs with AG capability and then unlocks their aesthetic cropping power via CoT-based rationales.
\item[$\bullet$] Extensive experiments show that Venus achieves SOTA performance in both AG and aesthetic cropping, enabling \textbf{interpretable} and \textbf{interactive} aesthetic refinement.
\end{itemize}

\section{Related Work}
\label{sec:related word}

\subsection{Image Aesthetic Tasks and Datasets}
Image aesthetic tasks can be roughly divided into seven levels~\cite{jin2024apddv2}, as shown in \cref{Fig.2}.
UNIAA~\cite{zhou2024uniaa} includes 27K images with annotations on aesthetic perception, descriptions, and scores. AesMMIT~\cite{huang2024Aesexpert} provides 88K human feedback on 21K images, supplemented by instructions generated by GPT-4~\cite{achiam2023gpt-4}. Qi \etal~\cite{qi2025PhotoEye} construct PhotoCritique by scraping and processing user comments from online platforms, resulting in 450K images paired with aesthetic descriptions. While these datasets provide valuable resources for research, they primarily focus on tasks such as aesthetic perception and description, with most evaluations being overly positive. As a result, models trained on them often struggle to effectively support the AG task.

\subsection{Aesthetic MLLMs}
Recent progress of aesthetic MLLMs has been made in tasks such as aesthetic perception and description, with models including UNIAA~\cite{zhou2024uniaa}, AesExpert~\cite{huang2024Aesexpert}, and PhotoEye~\cite{qi2025PhotoEye} achieving improved performance through LLaVA-based instruction fine-tuning~\cite{liu2024llava-v1.5}. In addition, Q-ALIGN~\cite{wu2023Q-align}, CALM~\cite{liu2025CALM}, and ArtiMuse~\cite{cao2025artimuse} further advance aesthetic scoring. However, as shown in \cref{Fig.1}, current MLLMs remain limited in addressing the more complex AG task and perform poorly in aesthetic cropping.

\subsection{Specialized Aesthetic Cropping Models}
\label{Section 2.3}
Aesthetic cropping enhances the visual appeal by re-framing images and is widely used in tasks such as thumbnail generation~\cite{suh2003suoluetu-1} and camera viewpoint recommendation~\cite{li2018A2-RL}. Existing methods fall into two main categories: (1) Anchor-based methods~\cite{zeng2019GAICD,su2024UGC,zhang2025procrop}, which score predefined cropping candidates and select the highest-scoring region; and (2) Regression-based methods~\cite{li2018A2-RL,hong2021CACNet,hong2024Gencrop}, which directly predict the optimal cropping coordinates. However, these models remain small-scale, architecturally diverse, and limited to the cropping task. They also lack interpretability and interactivity, which limits their adaptation to diverse and personalized aesthetic preferences.
\begin{figure*}[t]
\centering
\includegraphics[width=\textwidth]{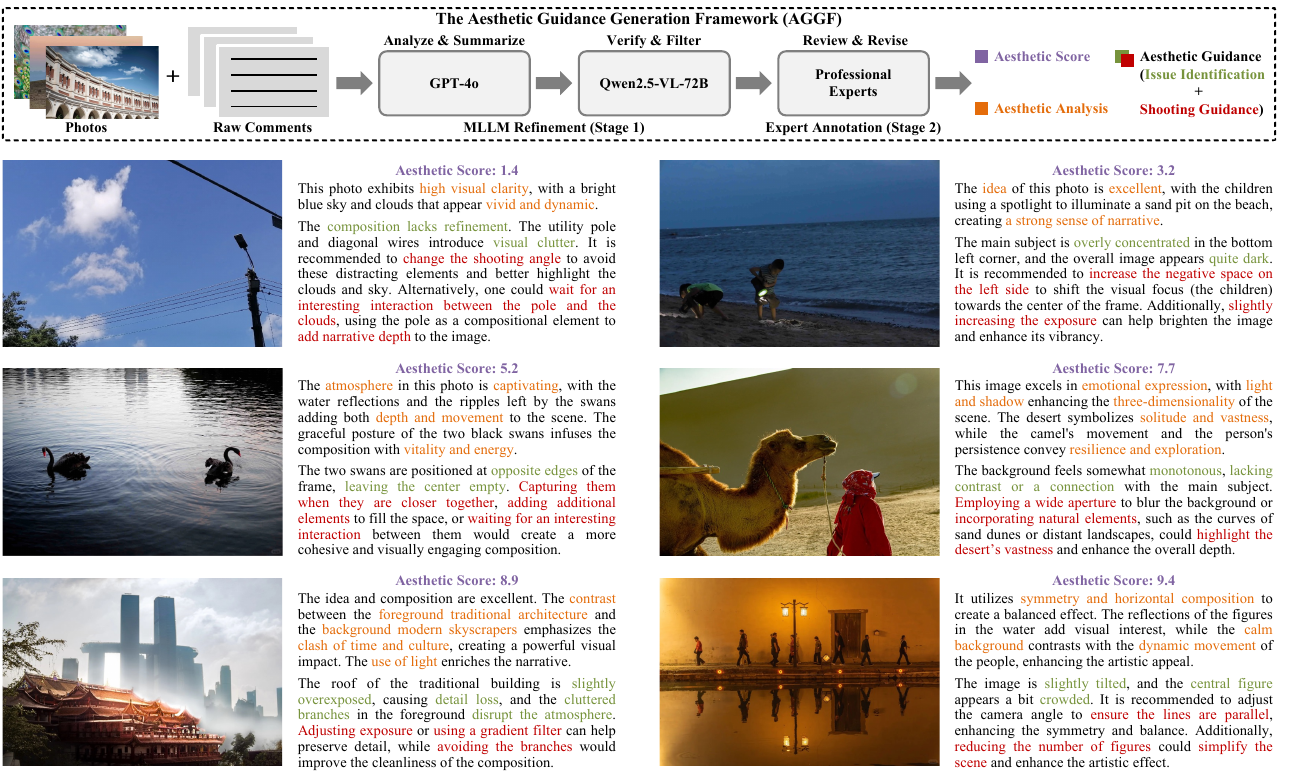}
\caption{Illustration of the AGGF and examples from the proposed AesGuide dataset, showing aesthetic scores (in purple), aesthetic analyses (in orange), and aesthetic guidance (issue identification in green and shooting guidance in red) on the right side of each image.}
\label{Fig.3}
\end{figure*}

\section{A New Dataset and Benchmark: AesGuide}
To the best of our knowledge, \textbf{AesGuide} is the first dataset and benchmark specifically designed for the AG task. It contains 10,748 real-world photos, each annotated with an aesthetic score, analysis, and guidance, along with three evaluation metrics and expert assessments for comprehensive benchmarking. AesGuide has two key features: \textbf{(1) Issue Identification.} Each photo is paired with professional critiques identifying aesthetic issues such as poor composition, lighting imbalance, or exposure errors. \textbf{(2) Shooting Guidance.} Each photo is further accompanied by actionable guidance for shooting or post-editing.

\subsection{Data Collection and Annotation}
The dataset is collected from two main sources:
(1) Crawling photos from multiple online platforms together with constructive comments from photographers and enthusiasts.
(2) Collaborating with professional photographers and studios to purchase photos for research use, each accompanied by at least one comment describing aesthetic flaws and improvement suggestions.
However, these raw comments are often noisy and inconsistent, making them unsuitable for direct supervised fine-tuning.

To address this, we propose the Aesthetic Guidance Generation Framework (AGGF), which integrates MLLM refinement and expert annotation to produce high-quality AG labels, as illustrated in \cref{Fig.3}. AGGF consists of two stages: \textbf{(1) MLLM Refinement.} Several MLLMs are employed to analyze and summarize the raw comments. \textbf{(2) Expert Annotation.} Professional photography experts review and revise the outputs to ensure accuracy and completeness. We invite 20 experts, including experienced freelance photographers, studio professionals, popular photography bloggers, and trained data annotators, ensuring both annotation quality and diversity. All experts are compensated and complete two rounds of aesthetic training before annotation.

Specifically, in the MLLM refinement stage, both the photo and its raw comments are fed into GPT-4o~\cite{hurst2024gpt-4o}, which is prompted to analyze and summarize the comments into an aesthetic critique. A smaller MLLM, Qwen2.5-VL-72B~\cite{bai2025qwen25}, then verifies whether the critique includes both issue identification and shooting guidance, filtering out incomplete cases. In the expert annotation stage, human experts review and revise the generated critiques, organizing the content into two parts: aesthetic analysis and aesthetic guidance. The former emphasizes interpreting and appreciating the photo from an aesthetic perspective, while the latter focuses on issue identification and shooting guidance. The dataset is divided into 20 annotation sessions, each assigned to one expert. For quality control, experts rate each photo on a 1–10 scale. If any individual score deviates by more than three points from the mean of the other experts, the photo is flagged for group discussion and re-annotation. An overview of the AGGF and examples from the proposed AesGuide dataset are shown in \cref{Fig.3}.

\subsection{Benchmark}
\label{Section 3.2}
We randomly sample 1,000 photos in AesGuide to construct the benchmark for the AG task. The remaining 9,748 photos are used for training. To ensure benchmark reliability, each annotation undergoes an additional round of cross-review by the other experts. Any inconsistencies are carefully examined through collective discussion, and revisions are made iteratively until full consensus is reached. This additional verification minimizes individual bias and ensures that all annotations reflect broad expert consensus and mainstream aesthetic standards.

Existing aesthetic benchmarks for MLLMs, such as AesBench~\cite{huang2024aesbench} and PhotoBench~\cite{qi2025PhotoEye}, primarily adopt multiple-choice questions evaluated by accuracy. However, such a design fails to reflect real-world scenarios, where users rarely provide predefined answer options. In practice, models are expected to autonomously generate descriptive feedback rather than select from fixed choices. Moreover, these benchmarks lack interpretability, making them inadequate for assessing the complex AG task. To address these limitations, we draw inspiration from recent studies~\cite{zheng2023gpt-useful-1, wu2023Qbench} showing that single-modal GPT can serve as a reliable evaluator for language-based tasks. For instance, Q-Bench~\cite{wu2023Qbench} reports a strong correlation between GPT-generated scores and human ratings, with SRCC consistently above 0.95.

\begin{figure*}[htbp]
\centering
\includegraphics[width=\textwidth]{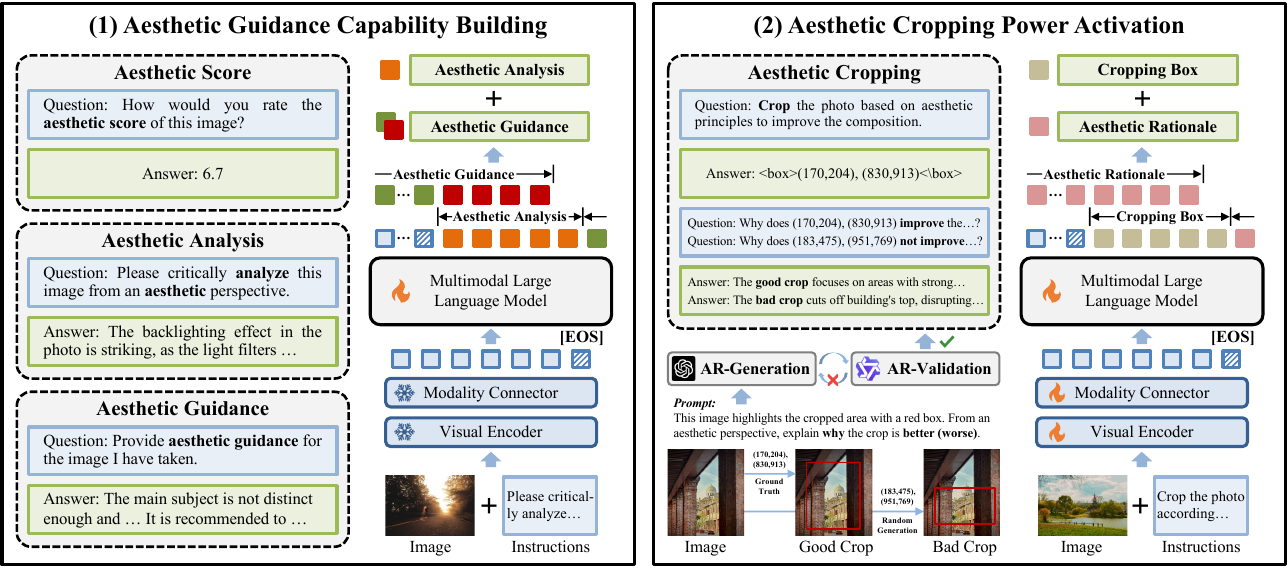}
\caption{Overview of the \textbf{Venus} framework: (1) Aesthetic guidance capability building, where AesGuide is leveraged to empower MLLMs with AG capability. (2) Aesthetic cropping power activation, which unlocks the cropping ability using CoT-based rationales.}
\label{Fig.4}
\end{figure*}

Building upon this evidence, we follow the setup of prior works~\cite{wu2023Qbench, wu2024Q-instruct, zhou2024uniaa} and adopt a single-modal GPT to evaluate the quality of MLLM outputs across three key dimensions:
\textbf{(1) Completeness.} Outputs are rewarded for covering more information consistent with the golden annotations.
\textbf{(2) Preciseness.} Outputs containing inaccurate or conflicting details compared with the golden annotations are penalized.
\textbf{(3) Relevance.} Outputs are encouraged to focus on issue identification and shooting guidance rather than irrelevant content.
Each dimension is scored on a three-point scale $[0,1,2]$.
For example, the prompt template for the Completeness evaluation is shown below:

\begin{tcolorbox}
    [
     colback=gray!10,  
     colframe=black,  
     leftrule={0.6pt},  
     rightrule={0.6pt},  
     toprule={0.6pt},  
     bottomrule={0.6pt},  
     left={2pt},  
     right={2pt},  
     top={1pt},  
     bottom={1pt}  
    ]
\small

\emph{\textbf{Prompt:} Evaluate whether [RESPONSE] contains the image's shortcomings or suggestions for improving the image mentioned in [ANSWER].}
\begin{enumerate}
    \item \emph{If [RESPONSE] does not contain the image's shortcomings or suggestions for improving the image mentioned in [ANSWER], rate a score of 0.}
    \item \emph{If [RESPONSE] contains some of the image's shortcomings or suggestions mentioned in [ANSWER], rate a score of 1.}
    \item \emph{If [RESPONSE] contains most of the image's shortcomings or suggestions mentioned in [ANSWER], rate a score of 2.}
\end{enumerate}
\end{tcolorbox}

\textbf{Expert Assessments.} To further ensure benchmark reliability and validate the soundness of the GPT-assisted evaluation for the AG task, we conduct additional expert assessments. 10 professionals, drawn from the original pool of 20 experts, evaluate 100 randomly sampled photos from the AesGuide benchmark. Each expert receives the same prompts used in the GPT-assisted evaluation and scores the MLLM outputs using an identical three-point scale $[0,1,2]$ based on Completeness, Preciseness, and Relevance. The final scores are obtained by averaging the ratings across all experts and evaluation dimensions.
\begin{table*}
  \caption{Evaluation results on the AesGuide benchmark. \textbf{Mean} denotes the average across three dimensions. \textbf{Expert} represents expert assessments, and \textbf{Rank} shows the order by Mean / Expert scores. Models are grouped by category: proprietary, aesthetic-oriented, and open-source general MLLMs. Best and second-best results are in \textbf{bold} and \underline{underlined}, respectively.}
  \centering
  \small
  \renewcommand{\arraystretch}{1.0}
  \setlength{\tabcolsep}{3.3mm}
  \begin{tabular}{l|c|cccc|c|c}
    \toprule
    \textbf{Model} & \textbf{Params} & \textbf{Completeness} & \textbf{Preciseness} & \textbf{Relevance} & \textbf{Mean} & \textbf{Expert} & \textbf{Rank} \\
    
    \midrule
    \rowcolor{gray!5}
    \multicolumn{8}{c}{\textbf{Proprietary MLLMs}}\\
    \midrule
    
    GPT-4o~\cite{hurst2024gpt-4o} & NA & 0.84 & 1.09 & 1.01 & 0.98 & 1.15 & 8 / 7 \\
    Gemini-2.0-Pro~\cite{team2023gemini} & NA & 1.09 & 1.12 & 1.36 & 1.19 & 1.16 & 6 / 6 \\
    Qwen-VL-Max~\cite{bai2023qwen-vl-max} & NA & 0.90 & 1.05 & 0.56 & 0.84 & 0.89 &  11 / 11 \\
    
    \midrule
    \rowcolor{gray!5}
    \multicolumn{8}{c}{\textbf{Aesthetic MLLMs}}\\
    \midrule
    
    AesExpert~\cite{huang2024Aesexpert} & 7B & 0.33 & 0.56 & 0.51 & 0.47 & 0.56 & 15 / 14 \\
    UNIAA~\cite{zhou2024uniaa} & 7B & 1.03 & 1.02 & 1.23 & 1.09 & 1.01 & 7 / 8 \\

    \midrule
    \rowcolor{gray!5}
    \multicolumn{8}{c}{\textbf{Open-source General MLLMs}}\\
    \midrule
    
    Qwen-VL-Chat~\cite{bai2023qwen-vl-chat} & 7B & 0.73 & 0.91 & 0.59 & 0.74 & 0.70 & 12 / 12 \\
    \textbf{Venus-Q (ours)} & 7B & 1.12 \textbf{\textcolor{popo_green}{\scriptsize +0.39}} & 1.23 \textbf{\textcolor{popo_green}{\scriptsize +0.32}} & 1.57 \textbf{\textcolor{popo_green}{\scriptsize +0.98}} & 1.31 \textbf{\textcolor{popo_green}{\scriptsize +0.57}} & 1.36 \textbf{\textcolor{popo_green}{\scriptsize +0.66}} & 5 / 4 \\ 
    \midrule
    InternVL 2.5~\cite{chen2024internVL2.5} & 7B & 0.83 & 1.01 & 1.02 & 0.95 & 0.99 & 10 / 9 \\
    \textbf{Venus-I (ours)} & 7B & \underline{1.27} \textbf{\textcolor{popo_green}{\scriptsize +0.44}} & \underline{1.33} \textbf{\textcolor{popo_green}{\scriptsize +0.32}} & \underline{1.81} \textbf{\textcolor{popo_green}{\scriptsize +0.79}} & \underline{1.47} \textbf{\textcolor{popo_green}{\scriptsize +0.52}} & \underline{1.50} \textbf{\textcolor{popo_green}{\scriptsize +0.51}} & \underline{2} / \underline{2} \\
    \midrule
    MiniCPM-V 2.6~\cite{yao2024minicpm} & 7B & 0.83 & 1.04 & 1.04 & 0.97 & 0.92 & 9 / 10 \\
    \textbf{Venus-M (ours)} & 7B & 1.19 \textbf{\textcolor{popo_green}{\scriptsize +0.36}} & 1.24 \textbf{\textcolor{popo_green}{\scriptsize +0.20}} & 1.72 \textbf{\textcolor{popo_green}{\scriptsize +0.68}} & 1.38 \textbf{\textcolor{popo_green}{\scriptsize +0.41}} & 1.30 \textbf{\textcolor{popo_green}{\scriptsize +0.38}} & 4 / 5 \\
    \midrule
    LLaVA-1.5-7B~\cite{liu2024llava-v1.5} & 7B & 0.64 & 0.79 & 0.35 & 0.59 & 0.52 & 14 / 15 \\
    \textbf{Venus-L-7B (ours)} & 7B & 1.26 \textbf{\textcolor{popo_green}{\scriptsize +0.62}} & 1.32 \textbf{\textcolor{popo_green}{\scriptsize +0.53}} & 1.80 \textbf{\textcolor{popo_green}{\scriptsize +1.45}} & 1.46 \textbf{\textcolor{popo_green}{\scriptsize +0.87}} & 1.40 \textbf{\textcolor{popo_green}{\scriptsize +0.88}} & 3 / 3 \\
    \midrule
    LLaVA-1.5-13B~\cite{liu2024llava-v1.5} & 13B & 0.67 & 0.86 & 0.41 & 0.65 & 0.61 & 13 / 13 \\
    \textbf{Venus-L-13B (ours)} & 13B & \textbf{1.28} \textbf{\textcolor{popo_green}{\scriptsize +0.61}} & \textbf{1.35} \textbf{\textcolor{popo_green}{\scriptsize +0.49}} & \textbf{1.83} \textbf{\textcolor{popo_green}{\scriptsize +1.42}} & \textbf{1.49} \textbf{\textcolor{popo_green}{\scriptsize +0.84}} & \textbf{1.53} \textbf{\textcolor{popo_green}{\scriptsize +0.92}} & \textbf{1} / \textbf{1} \\
    \bottomrule
  \end{tabular}
  \label{Tab.2}
\end{table*}

\section{Method}
In the following sections, we introduce the Venus framework, as illustrated in \cref{Fig.4}. Venus comprises two stages: \textbf{(1) Aesthetic guidance capability building}, where AesGuide is leveraged to empower MLLMs with AG capability, resulting in aesthetic guidance MLLMs. \textbf{(2) Aesthetic cropping power activation}, which unlocks the cropping ability of the aesthetic guidance MLLM using CoT-based rationales. Experiments show that embedding AG capability within the MLLM is essential for aesthetic cropping.

\subsection{Aesthetic Guidance Capability Building}
\label{Section 4.1}
The training of open-source MLLMs typically follows a two-step pipeline. In the first step, million-scale web data are used to align the representation spaces of the visual encoder and the large language model (LLM)~\cite{sharma2018instruction-stage1-2, schuhmann2021instruction-stage1-1}. In the second step, supervised instruction fine-tuning~\cite{liu2023llava} is performed on human-labeled datasets~\cite{yu2016instruction-stage2-2, marino2019instruction-stage2-1}.
Building upon this pipeline, we apply supervised instruction fine-tuning on the AesGuide dataset to equip the MLLM with AG capabilities. During this process, the visual encoder and modality connector are frozen, while only the LLM is updated.

To build AG capability, the MLLM is trained to progressively address increasingly complex aesthetic questions, as shown in \cref{Fig.4} (1).
This design draws on the human aesthetic reasoning process, which evolves from forming an overall impression to analyzing visual strengths and weaknesses, and finally to proposing actionable refinements.
Through this progression, the model learns not only to perform aesthetic analysis but also to identify the causes of visual issues and suggest meaningful improvements, fostering a more human-like understanding of aesthetics.
Each data sample is represented as \( (x, q, a, g) \), where \( x \) is the input image, \( q \) is the instruction, \( a \) is the aesthetic analysis, and \( g \) is the aesthetic guidance. The training objective is to maximize the probability of generating \( a \) and \( g \) given \( (x, q) \):
\begin{equation}
\mathcal{L}_{AG} = -\mathbb{E}_{(x, q, a, g) \sim \mathcal{AG}} \sum\limits_{t=1}^{T} \log \pi_\theta(y_t \mid x, q, y_{<t}),
\end{equation}
where \( \mathcal{AG} \) represents the AesGuide dataset, \( y \) is the concatenated sequence of \( a \) and \( g \), and \( \pi_\theta \) denotes the token distribution of the MLLM. The resulting aesthetic guidance MLLM, \( \pi_{\text{AG}} \), serves as the initialization for the next stage, providing a solid foundation for aesthetic cropping.

We apply supervised instruction fine-tuning~\cite{liu2023llava} to five open-source general MLLMs with diverse architectures, including Qwen-VL-Chat~\cite{bai2023qwen-vl-chat}, InternVL 2.5~\cite{chen2024internVL2.5}, MiniCPM-V 2.6~\cite{yao2024minicpm}, LLaVA-1.5-7B~\cite{liu2024llava-v1.5}, and LLaVA-1.5-13B~\cite{liu2024llava-v1.5}. The resulting fine-tuned models are denoted as Venus-Q, Venus-I, Venus-M, Venus-L-7B, and Venus-L-13B, respectively. The architectural details of the five baseline MLLMs are summarized in the Appendix.

\subsection{Aesthetic Cropping Power Activation}
In the second stage of Venus, we activate the aesthetic cropping power of the aesthetic guidance MLLM using high-quality CoT-based rationales, as shown in \cref{Fig.4} (2). As discussed in \cref{Section 2.3}, existing cropping models do not learn the compositional logic behind cropping decisions---they simply score candidates or regress coordinates, without capturing the underlying aesthetic reasoning or visual intent.

To address this issue, we generate high-quality aesthetic rationales (AR) for each cropping box to guide the MLLM in reasoning about its cropping decisions.
CoT~\cite{chu2023CoT, xu2024llava-cot, jiang2025ArtCoT} enhances the reasoning ability of MLLMs by guiding them to produce explicit reasoning traces, which we find crucial for activating their aesthetic cropping power.
Unlike existing methods that only learn from labeled ``good crops,'' we also include randomly generated ``bad crops'' with suboptimal compositions (\eg, those that disrupt subject integrity or focus on irrelevant regions) and generate aesthetic rationales for both.
To ensure the reliability of these rationales, we design a two-stage process:
\textbf{(1) AR-Generation.} GPT-4o~\cite{hurst2024gpt-4o} is prompted with the original image, where the cropping region is outlined in red, and instructed to explain why the area inside the box exhibits better or worse composition.
\textbf{(2) AR-Validation.} The generated rationales are then reviewed by Qwen2.5-VL-72B~\cite{bai2025qwen25}, which verifies whether each explanation aligns with the visual content and correctly evaluates the cropping quality. Cases that fail validation or exhibit low-quality reasoning are automatically flagged for re-generation.
This cycle repeats until all aesthetic rationales meet the alignment and quality standards.

In this stage, the model is trained to jointly predict the cropping box and its rationale, grounding spatial decisions in aesthetic reasoning. 
The aesthetic guidance MLLM is fine-tuned in a full-schedule manner, with each data sample represented as \( (x, q, b, r) \), where \( x \) is the input image, \( q \) is the instruction, \( b \) is the cropping box, and \( r \) is the aesthetic rationale. The training objective is to maximize the conditional probability of generating \( b \) and \( r \) given \( (x, q) \):
\begin{equation}
\mathcal{L}_{AC} = -\mathbb{E}_{(x, q, b, r) \sim \mathcal{D}} \sum\limits_{t=1}^{T} \log \pi_{\text{AG}}(y_t \mid x, q, y_{<t}),
\end{equation}
where \( \mathcal{D} \) represents the aesthetic cropping dataset, \( y \) is the concatenated sequence of \( b \) and \( r \), and \( \pi_{\text{AG}} \) denotes the aesthetic guidance MLLM from Stage~1. We select Venus-Q to activate its aesthetic cropping power.

\begin{table}[t]
\caption{Quantitative results on the FLMS benchmark. \textbf{Inp.} and \textbf{Ina.} denote interpretability and interactivity, respectively. Best and second-best results are in \textbf{bold} and \underline{underlined}, respectively}
\centering
\small
\renewcommand{\arraystretch}{1}
\setlength{\tabcolsep}{1.1mm}
\begin{tabular}{l|ccc|c@{\hskip 0.9mm}c}
\toprule
\textbf{Model} & \textbf{IoU\%($\uparrow$)} & \textbf{Disp($\downarrow$)} & \textbf{R\%($\uparrow$)} &
\textbf{Inp.} & 
\textbf{Ina.}\\ 

\midrule
\rowcolor{gray!5}
\multicolumn{6}{c}{\textbf{Specialized Aesthetic Cropping Models}}\\
\midrule
ASM-Net~\cite{tu2020asm-net} & 84.86 & 0.0390 & - & \ding{56} & \ding{56} \\
CACNet~\cite{hong2021CACNet} & 85.40 & \underline{0.0330} & - & \ding{56} & \ding{56} \\
HCIC~\cite{zhang2022HCIC} & 85.00 & 0.0340 & - & \ding{56} & \ding{56} \\
SAC-Net~\cite{yang2023SAC-Net} & \underline{85.51} & 0.0333 & - & \ding{56} & \ding{56} \\
UNIC~\cite{liu2023UNIC} & 84.00 & 0.0370 & - & \ding{56} & \ding{56} \\
ProCrop~\cite{zhang2025procrop} & 84.30 & 0.0360 & - & \ding{56} & \ding{56} \\

\midrule
\rowcolor{gray!5}
\multicolumn{6}{c}{\textbf{Proprietary MLLMs}}\\
\midrule

GPT-4o~\cite{hurst2024gpt-4o} & 71.61 & 0.0615 & 43.2 & \checkmark & \checkmark \\
Gemini-2.0-Pro~\cite{team2023gemini} & 68.15 & 0.0647 & 38.4 & \checkmark & \checkmark \\
Qwen-VL-Max~\cite{bai2023qwen-vl-max} & 58.09 & 0.0814 & 20.4 & \checkmark & \checkmark \\

\midrule
\rowcolor{gray!5}
\multicolumn{6}{c}{\textbf{Aesthetic MLLMs}}\\
\midrule

AesExpert~\cite{huang2024Aesexpert} & 37.86 & 0.1206 & 9.6 & \checkmark & \checkmark \\
UNIAA~\cite{zhou2024uniaa} &  21.03 & 0.1602 & 0.2 & \checkmark & \checkmark \\

\midrule
\rowcolor{gray!5}
\multicolumn{6}{c}{\textbf{Open-source General MLLMs}}\\
\midrule

Qwen-VL-Chat~\cite{bai2023qwen-vl-chat} & 73.84 & 0.0667 & \underline{67.2} & \checkmark & \checkmark \\
Qwen2.5-VL-7B~\cite{bai2025qwen25} & 50.32 & 0.1056 & 23.6 & \checkmark & \checkmark \\
Qwen2.5-VL-32B~\cite{bai2025qwen25} & 54.35 & 0.0957 & 19.0 & \checkmark & \checkmark \\
InternVL 2.5~\cite{chen2024internVL2.5} & 71.53 & 0.0705 & 51.2 & \checkmark & \checkmark \\
MiniCPM-V 2.5~\cite{yao2024minicpm} & 49.29 & 0.1158 & 12.5 & \checkmark & \checkmark \\
MiniCPM-V 2.6~\cite{yao2024minicpm} & 45.95 & 0.1111 & 5.2 & \checkmark & \checkmark \\
LLaVA-1.5-7B~\cite{liu2024llava-v1.5} & 54.30 & 0.0793 & 21.6 & \checkmark & \checkmark \\
LLaVA-1.5-13B~\cite{liu2024llava-v1.5} & 59.04 & 0.0876 & 32.0 & \checkmark & \checkmark \\

\midrule
\multirow{2}{*}{\textbf{Venus-Q (ours)}} & \textbf{87.01} & \textbf{0.0292} & \textbf{92.0} & \multirow{2}{*}{\checkmark} & \multirow{2}{*}{\checkmark} \\
& \textbf{\textcolor{popo_green}{+1.50}} & \textbf{\textcolor{popo_green}{-0.0038}} & \textbf{\textcolor{popo_green}{+24.8}} & & \\
\bottomrule
\end{tabular}
\label{Tab.3}
\end{table}

\section{Experiments}
\subsection{Implementation Details}
\textbf{Datasets.} In Stage 1, we conduct experiments on AesGuide, which serves as a pioneering benchmark for the AG task; following UNIAA~\cite{zhou2024uniaa}, we also leverage a few open-source datasets~\cite{murray2012AVA,chang2017PCCD, lee2018KU-PCP,yang2022PARA} as auxiliary training data. In Stage 2, we train primarily on three public cropping datasets, including CPC~\cite{wei2018CPC=VEN}, GAIC~\cite{zeng2020GAICE}, and FCDB~\cite{chen2017FCDB}, and additionally include a small expert-annotated set as a supplement to the public annotations.
We evaluate on the widely used FLMS benchmark~\cite{fang2014FLMS} for aesthetic cropping to assess out-of-domain generalization. 

\noindent \textbf{Evaluation Metrics.} Following prior works~\cite{fang2014FLMS, zhang2022HCIC}, we use Intersection-over-Union (IoU) and boundary displacement (Disp) as the primary metrics for aesthetic cropping, and additionally report the recall rate (R), which measures the proportion of high-quality crops with IoU $\geq 75\%$, providing a complementary view of overall cropping quality.

\begin{figure*}[t]
\centering
\includegraphics[width=\textwidth]{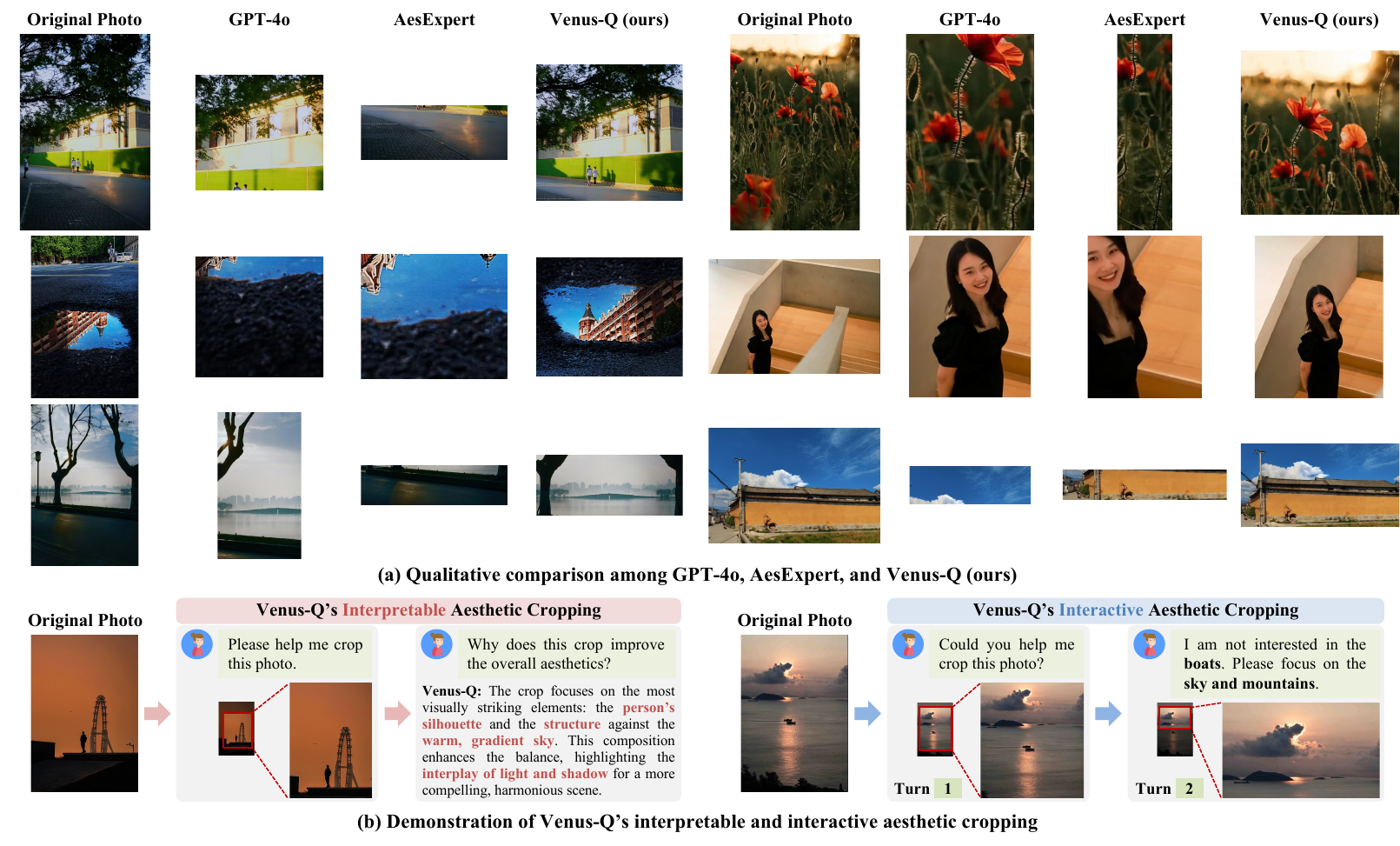}
\caption{Qualitative comparison of aesthetic cropping results among GPT-4o, AesExpert, and Venus-Q (ours) (a), along with a demonstration of Venus-Q’s \textbf{interpretable} and \textbf{interactive} aesthetic cropping capabilities (b).}
\label{Fig.5}
\end{figure*}

\noindent \textbf{Evaluated Methods.} In the AG evaluation on the AesGuide benchmark, we compare our five aesthetic guidance MLLMs with three proprietary MLLMs, including GPT-4o~\cite{hurst2024gpt-4o}, Gemini-2.0-Pro~\cite{team2023gemini}, and Qwen-VL-Max~\cite{bai2023qwen-vl-max}, as well as two open-source aesthetic-oriented ones, AesExpert~\cite{huang2024Aesexpert} and UNIAA~\cite{zhou2024uniaa}. For aesthetic cropping evaluation, in addition to the above MLLMs, we further compare Venus-Q with SOTA specialized aesthetic cropping models.

\noindent \textbf{Training Settings.} We train the MLLMs on 8 NVIDIA A40 GPUs (48 GB each), applying the LoRA technique~\cite{hu2022lora} in the first stage and full-parameter fine-tuning in the second. Each stage is trained for three epochs using the default hyperparameters of the respective MLLMs.

\begin{table}[t]
\caption{Ablation study on the AesGuide benchmark.}
\centering
\small
\renewcommand{\arraystretch}{1}
\setlength{\tabcolsep}{1.3mm}
\begin{tabular}{lcccc}
\toprule
\textbf{Settings} & \textbf{Com.} & \textbf{Pre.} & \textbf{Rel.} & \textbf{Mean} \\
\midrule
w/o Aesthetic Analysis in Stage 1 & 1.09 & 1.17 & 1.41 & 1.22 \\
\rowcolor{gray!15} \textbf{Venus-Q (ours)} & \textbf{1.12} & \textbf{1.23} & \textbf{1.57} & \textbf{1.31} \\
\bottomrule
\end{tabular}
\label{Tab.4}
\end{table}

\begin{table}[t]
\caption{Ablation study on the FLMS benchmark.}
\centering
\small
\renewcommand{\arraystretch}{1}
\setlength{\tabcolsep}{1.6mm}
\begin{tabular}{lccc}
\toprule
\textbf{Settings} & \textbf{IoU\%($\uparrow$)} & \textbf{Disp($\downarrow$)} & \textbf{R\%($\uparrow$)} \\
\midrule
w/o AesGuide in Stage 1 & 84.90 & 0.0327 & 86.2 \\
w/o AR-Generation in Stage 2 & 81.10 & 0.0402 & 75.2 \\
w/o AR-Validation in Stage 2 & 86.47 & 0.0308 & 90.4 \\
\rowcolor{gray!15} \textbf{Venus-Q (ours)} & \textbf{87.01} & \textbf{0.0292} & \textbf{92.0} \\
\bottomrule
\end{tabular}
\label{Tab.5}
\end{table}

\subsection{Main Results}
\noindent \textbf{Quantitative Analysis.} 
As illustrated in \cref{Tab.2}, all five aesthetic guidance MLLMs achieve consistent gains on the AesGuide benchmark, outperforming both proprietary and aesthetic-oriented MLLMs across all dimensions. 
The GPT-assisted rankings align closely with expert assessments, confirming GPT as a reliable tool for evaluating AG quality. 
In aesthetic cropping (\cref{Tab.3}), Venus-Q surpasses both proprietary and aesthetic MLLMs and further outperforms specialized cropping models, improving IoU by +1.50 on FLMS~\cite{fang2014FLMS} over the previous SOTA and by +15.4 compared with GPT-4o~\cite{hurst2024gpt-4o}. Since aesthetic judgment is subjective and shaped by diverse user preferences, there is no single correct answer, making it essential for models to balance accuracy with transparent reasoning and adaptive interaction.
However, existing approaches struggle to meet these demands: specialized cropping models achieve competitive accuracy but are opaque, whereas MLLMs are explainable yet imprecise. Venus-Q bridges this gap by integrating the strengths of both paradigms, combining strong performance with interpretability and interactivity.

\noindent \textbf{Qualitative Analyses.}
As shown in \cref{Fig.5}, we compare Venus-Q with GPT-4o~\cite{hurst2024gpt-4o} and AesExpert~\cite{huang2024Aesexpert} on aesthetic cropping, where Venus-Q consistently produces more balanced and aesthetically refined results. Built upon an AG-enhanced MLLM foundation, Venus-Q leverages rich aesthetic knowledge to identify distracting regions and infer compositional intent—much like a trained photographer refining framing with purpose. Beyond generating cropping boxes, it also provides interpretable aesthetic rationales that clearly explain its decisions and supports interactive refinement through natural language feedback (\eg, emphasizing subjects, adjusting balance, or shifting focal regions), enabling explainable and personalized aesthetic enhancement.

\subsection{Ablation Studies}
\textbf{Are the progressive aesthetic questions essential for AG learning?}
We train Qwen-VL-Chat~\cite{bai2023qwen-vl-chat} in Stage 1 of Venus using only the aesthetic guidance data. As shown in \cref{Tab.4}, removing the aesthetic analysis leads to a consistent drop across all metrics, confirming the necessity of the full progressive design for robust AG capability.

\noindent \textbf{Is AG capability crucial for aesthetic cropping?}
We exclude AesGuide from Stage 1 and perform Stage 2 of Venus as usual. As shown in \cref{Tab.5}, performance drops across all metrics on FLMS~\cite{fang2014FLMS}, confirming that AG pretraining is vital for establishing aesthetic understanding and that Venus relies on it to achieve superior cropping performance.

\noindent \textbf{Does the CoT-based rationale design in Stage 2 enhance aesthetic cropping performance?}
We conduct ablation studies on FLMS~\cite{fang2014FLMS} (\cref{Tab.5}) to evaluate the impact of aesthetic rationales.
When AR-Generation is removed, the model learns only from cropping boxes without explicit reasoning supervision, leading to a clear drop across all metrics.
Similarly, disabling AR-Validation also degrades performance, underscoring the importance of rationale reliability for stable learning.
These results confirm that both AR-Generation and AR-Validation are essential for achieving the superior aesthetic cropping performance of Venus-Q.
\section{Conclusion}
We define \textbf{aesthetic guidance (AG)} and introduce \textbf{AesGuide}, the first dataset and benchmark for this task. Building on it, we propose \textbf{Venus}, a two-stage framework that first empowers MLLMs with AG capability and then activates their aesthetic cropping power. Experiments show that Venus achieves SOTA performance in both AG and aesthetic cropping, enabling \textbf{interpretable} and \textbf{interactive} refinement across both stages of photo creation. This work moves computational aesthetics beyond passive assessment toward actionable, human-aligned visual improvement.

\section*{Ackowledgements}
This work was supported by the grants from the National Natural Science Foundation of China (62525201, 62132001, 62432001) and Beijing Natural Science Foundation (L247006).

{
    \small
    \bibliographystyle{ieeenat_fullname}
    \bibliography{main}
}


\end{document}